\documentclass[10pt,twocolumn,letterpaper]{article}

\usepackage{wacv}              

\usepackage{graphicx}
\usepackage{amsmath}
\usepackage{amssymb}
\usepackage{booktabs}
\usepackage{multirow}

\usepackage[pagebackref,breaklinks,colorlinks]{hyperref}

\usepackage[capitalize]{cleveref}
\crefname{section}{Sec.}{Secs.}
\Crefname{section}{Section}{Sections}
\Crefname{table}{Table}{Tables}
\crefname{table}{Tab.}{Tabs.}

\def\wacvPaperID{1034} 
\def\confName{WACV}
\def\confYear{2024}

\begin{document}

\title{RGB-D Mapping and Tracking in a Plenoxel Radiance Field}


\author{
Andreas L. Teigen\textsuperscript{1}\thanks{Authors contributed equally to this work.} \and Yeonsoo Park\textsuperscript{2}\footnotemark[1] 
\and Annette Stahl\textsuperscript{1} \and Rudolf Mester\textsuperscript{1}   \and \\
\textsuperscript{1}Norwegian University of Science and Technology (NTNU), Trondheim, Norway\\
\textsuperscript{2}Mobiltech, Seoul, Republic of Korea\\
\hfill{\tt\small andreas.l.teigen@ntnu.no}\hfill{\tt\small yspark@mobiltech.io}\hfill
}
\maketitle


\begin{abstract}
   The widespread adoption of Neural Radiance Fields (NeRFs) have ensured significant advances in the domain of novel view synthesis in recent years. These models capture a volumetric radiance field of a scene, creating highly convincing, dense, photorealistic models through the use of simple, differentiable rendering equations. Despite their popularity, these algorithms suffer from severe ambiguities in visual data inherent to the RGB sensor, which means that although images generated with view synthesis can visually appear very believable, the underlying 3D model will often be wrong.  This considerably limits the usefulness of these models in practical applications like Robotics and Extended Reality (XR), where an accurate dense 3D reconstruction otherwise would be of significant value. In this paper, we present the vital differences between view synthesis models and 3D reconstruction models. We also comment on why a depth sensor is essential for modeling accurate geometry in general outward-facing scenes using the current paradigm of novel view synthesis methods. Focusing on the structure-from-motion task, we practically demonstrate this need by extending the Plenoxel radiance field model: Presenting an analytical differential approach for dense mapping and tracking with radiance fields based on RGB-D data without a neural network. Our method achieves state-of-the-art results in both mapping and tracking tasks, while also being faster than competing neural network-based approaches. The code is available at: \url{https://github.com/ysus33/RGB-D_Plenoxel_Mapping_Tracking.git}.
\end{abstract}

\begin{figure}[t]
  \centering
   \includegraphics[width=1\linewidth]{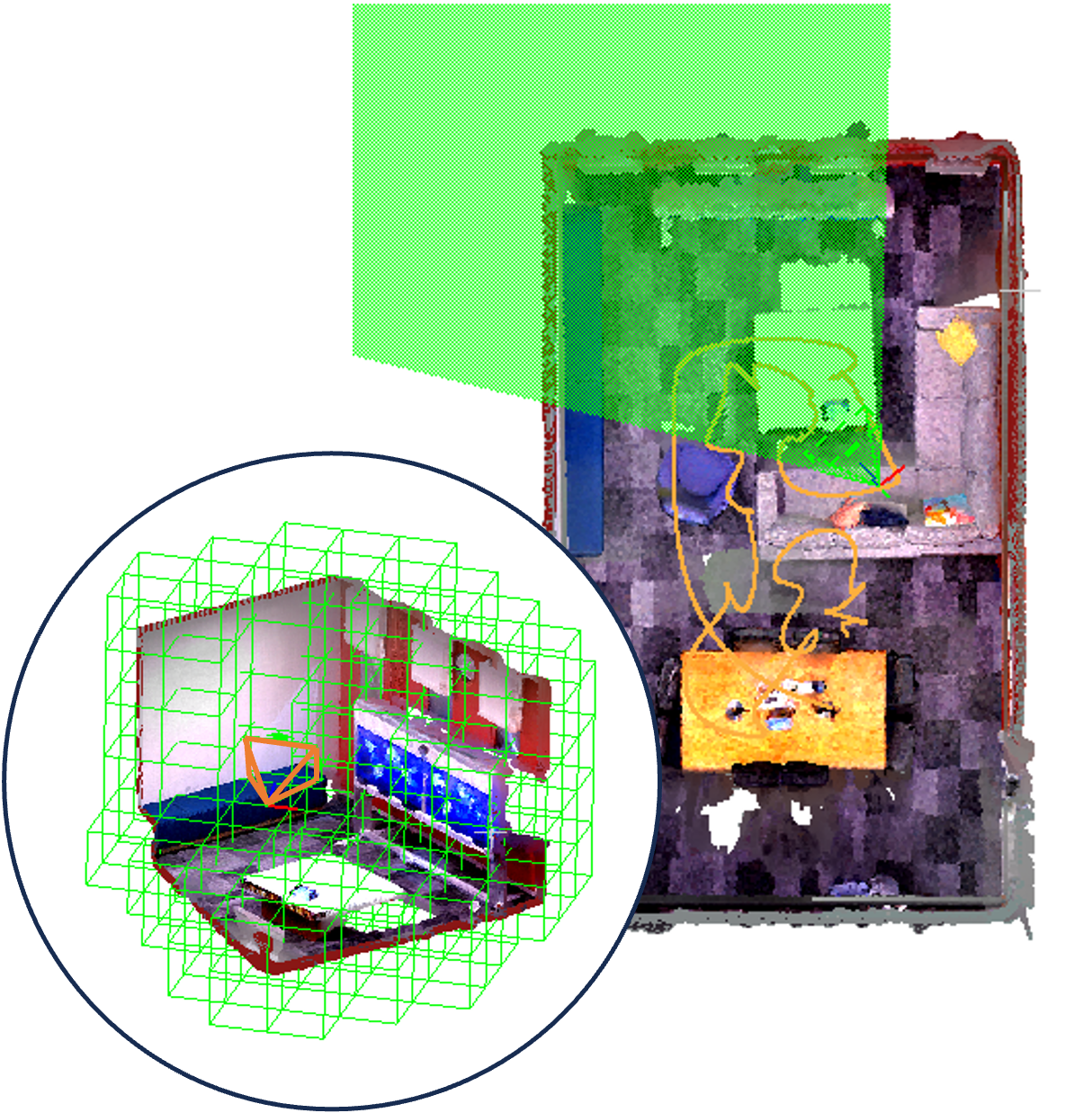}
   \caption{Visualization of the generated map and estimated trajectory on Office-3 of Replica dataset. The figure illustrates the voxel grid radiance field of the map.}
   \label{fig:door}
\end{figure}

\section{Introduction}
\label{sec:intro}


In the computer vision field, a dense map can be defined as a continuous 3D surface map generated using all observed pixels in an image set. The RGB-D sensor is a popular sensor choice for creating dense maps due to its rich geometric and photometric information. Dense maps are very useful for many tasks, such as path planning and collision avoidance in robotics, the interaction between real-world geometry and digital objects in \emph{extended reality} (XR), and simply as maps for human inspection. However, dealing with the prevalence of noise and missing measurements in the sensor data and the sheer amount of data collected can be challenging. 
In many cases, efficient processing of such data may require the utilization of a GPU. Dense maps can also be created from only RGB images, but this requires some assumption regarding areas of the scene that contain no gradients. 
Although dense mapping generally requires more effort than sparse, feature-based mapping, they contain much more information, and tracking the camera motion in a known environment with a good dense map is highly accurate as all image and model information can be used. This is opposed to sparse tracking, which only uses a subset of the available information, usually in the form of easily recognizable points in the images/model. Therefore, dense tracking is positioned to result in a smoother and more accurate trajectory.

The learned radiance field algorithms, such as \emph{Neural Radiance Fields} (NeRF) \cite{mildenhall2021nerf}, were introduced as a mathematically simple formulation for creating photorealistic dense models using RGB images. This works very well for some types of scenes, especially object-centric and inward-facing scenes and scenes with significant gradient coverage. These kinds of scenes significantly reduce the need to make assumptions in the modeling stage, resulting in highly accurate photogrammetry models.
However, if you apply these algorithms on outward-facing scenes and/or scenes with sparse gradient coverage, they will be incapable of capturing the geometry, and the view synthesis will not be valid outside of the camera pose distribution that was used for the training of the model.
Imbuing the NeRF optimization scheme to allow for full use of the RGB-D sensor removes ambiguities in gradient-less regions of the scene and makes it more useful for practical applications. 

We base our work on the Plenoxel algorithm \cite{fridovich2022plenoxels}: The analytical radiance field representation that does not use a neural network but rather a voxel grid representation. Building upon this representation, we make it more applicable for practical tasks by augmenting the algorithm to incorporate depth data. We then implement a pose optimization algorithm to track a camera throughout the scene, using volumetric, dense image-to-model alignment based on the radiance field rendering equations. We show the analytical derivations of all equations used for both optimization strategies and implement them in CUDA for fast computation times. Leveraging the inherent speed of the Plenoxel model of more than two magnitudes faster than the original NeRF algorithm, we create a very efficient mapping algorithm and a real-time, highly accurate tracking algorithm.

Our contributions are summed up as follows:
\begin{itemize}
    \item Present a discussion on the differences between models for novel view synthesis and models for 3D reconstruction and why this might lead to problems directly attempting to use NeRF for many practical applications like robotics and XR.
    \item Derive the analytical derivative equations for mapping and tracking in a voxel-based radiance field based on RGB-D data for efficient optimization in CUDA. 
    \item Showing improvement in both mapping and tracking results compared to existing radiance field mapping and tracking methods given the same time constraints. 
\end{itemize}

\section{Related Work}

\subsection{Dense Mapping and Tracking}
Despite the usefulness of dense visual mapping and the accuracy of dense tracking \cite{forster2016svo}, dense mapping and tracking methods \cite{newcombe2011dtam, izadi2011kinectfusion} have received relatively little attention compared to their sparse counterparts\cite{mur2015orb, engel2017direct}. This is mostly due to their technical and computational complexity as well as their reliance on either a depth sensor or an assumption for image regions with no gradients. Despite this, there have still been several noteworthy papers on the topic in the past few years: The DTAM algorithm \cite{newcombe2011dtam} by Newcombe et al. first proposed the idea of performing dense \emph{simultaneous localization and mapping} (SLAM) by separating the problem into alternating the tasks of updating a dense 3D model and tracking the camera pose by aligning the camera image to the model using randomly sampled image pixels. 
Kinect-fusion \cite{izadi2011kinectfusion} built on the premise , but
distinguished itself by relying only on a depth-only sensor and representing the entire model as a truncated signed distance field, using the iterative closest point algorithm for pose optimization. 
A more recent RGB-D SLAM paper: Bad-SLAM \cite{schops2019bad} creates a dense map based on surfels instead of individual pixels. The surfels' position, orientation, and size are optimized with a clever bundle adjustment implementation. 

The popularity of dense mapping and tracking has increased through the use of deep learning-based methods \cite{bloesch2018codeslam, teed2021droid}, often by the use of pixel-level depth estimators. CodeSLAM \cite{bloesch2018codeslam} does this by training a variational auto-encoder offline and estimating pixel depth based on RGB images. DROID-SLAM \cite{teed2021droid} is a surprisingly robust method compared to other pre-trained deep learning-based SLAM algorithms, providing good results on several different datasets, even for some datasets not included in the training set.

\subsection{Radiance Fields}
\emph{Neural Radiance Fields} (NeRF) \cite{mildenhall2021nerf} is a new technology that has taken the research community by storm. It allows for the creation of photo-realistic, dense, volumetric 3D view synthesis models of real-world objects, only requiring posed RGB images of the scene. NeRF natively stores a model in a highly compressed format as a neural network. The model is trained using ray-based volumetric rendering functions for training on images using multi-view consistency to produce a globally consistent model.
Ever since NeRF was proposed, there has been a lot of research to expand its applicability: \\

\noindent \textbf{Speed and efficiency.} 
Despite the original NeRF's remarkable capabilities, early versions were hindered by limitations of slow convergence rates. Numerous strategies\cite{barron2022mip,zhang2022fast, reiser2021kilonerf} have been proposed to improve its efficiency for both training and rendering.
Several papers \cite{muller2022instant, fridovich2022plenoxels} also show that it is possible to improve the speed by several orders of magnitude by changing the model representation. Muller et al. \cite{muller2022instant} do this by utilizing a hash encoding that enables disambiguation of hash collisions and consequently allows a smaller neural network to represent a larger scene. 
Fridovich et al. \cite{fridovich2022plenoxels} completely discard the neural network, favoring a voxel grid where each vertex is modeled by 28 float values, with samples at arbitrary locations are retrieved by linear interpolation of the eight closest voxels. Direct data access greatly improves the efficiency at the cost of increased memory consumption.
\\

\noindent \textbf{Depth guided NeRF.}
One direction of NeRF research has been to incorporate depth sensors to aid convergence speed \cite{deng2022depth} and geometric accuracy \cite{dey2022mip} of the NeRF model. The optimization based on the depth data uses the same underlying rendering equations as the color-only optimization and integrates seamlessly with the base model.\\

\noindent \textbf{Pose optimization.} Wang et al.\cite{wang2021nerf} have demonstrated the possibility of optimizing intrinsic camera parameters along with the neural radiance fields.
This was followed by Yen et al. \cite{yen2021inerf}, who proposed camera pose optimization based on a trained NeRF model, using image-to-model alignment.
Lin et al. \cite{lin2021barf} went further and showed that given a coarse initialization of the camera poses, the pose optimization for all training images could be done simultaneously during the training of the model.
However, to improve both the radius of convergence and performance, gradually, more layers of frequency encoding had to be introduced during the optimization process, which required human supervision due to scene-specific variations.\\

\noindent \textbf{Mapping and tracking.} Due to NeRF's simple formulation of dense mapping and its small storage size, several authors have attempted to use NeRF as a map representation in dense SLAM algorithms \cite{sucar2021imap, zhu2022nice, yang2022vox}. 
The first to attempt this for real-time processing was Sucar et al. \cite{sucar2021imap}, who used the original NeRF model \cite{mildenhall2021nerf} and an RGB-D sensor both for speed and to solve the geometric ambiguity problem covered in \cref{sec:nerfs_for_xr_and_robotics}. Although NeRFs are slow to train, the rapid convergence speed in the earlier epochs is a great help in achieving passable real-time performance. 
They showed promising results in a room-scale scene, but an increase in scene size would lead to catastrophic forgetting \cite{goodfellow2013empirical}.
Zhu et al. \cite{zhu2022nice} addressed the weakness of catastrophic forgetting by storing values from the learned model in a grid, which then would be queried by pre-trained decoder networks. 
Their follow-up work in \cite{zhu2023nicer} continued this trend of offloading tasks to pre-trained networks, replacing the depth sensor with two convolutional networks predicting a depth map and a normal map, respectively.
Vox-Fusion\cite{yang2022vox} adopts a similar concept to \cite{sucar2021imap} but achieves significantly reduced memory consumption by dynamically allocating sparse voxels based on an octree structure. 
\\

\noindent \textbf{NeRF as visual SLAM backend.} Notable but less relevant papers also use NeRF models as a visually pleasing back-end on existing sparse SLAM algorithms \cite{chung2022orbeez, rosinol2022nerf}. Several easy-to-use NeRF libraries \cite{tancik2023nerfstudio, muller2022instant} use such an approach, using methods such as \cite{schoenberger2016sfm} as a separate front-end to calculate camera poses that are then used as input for the NeRF model. We use an entirely different method that elegantly utilizes the
simple rendering equations of NeRF to optimize both the camera pose and the map with the same equation.



\begin{figure}[t]
  \centering
   \includegraphics[width=1\linewidth]{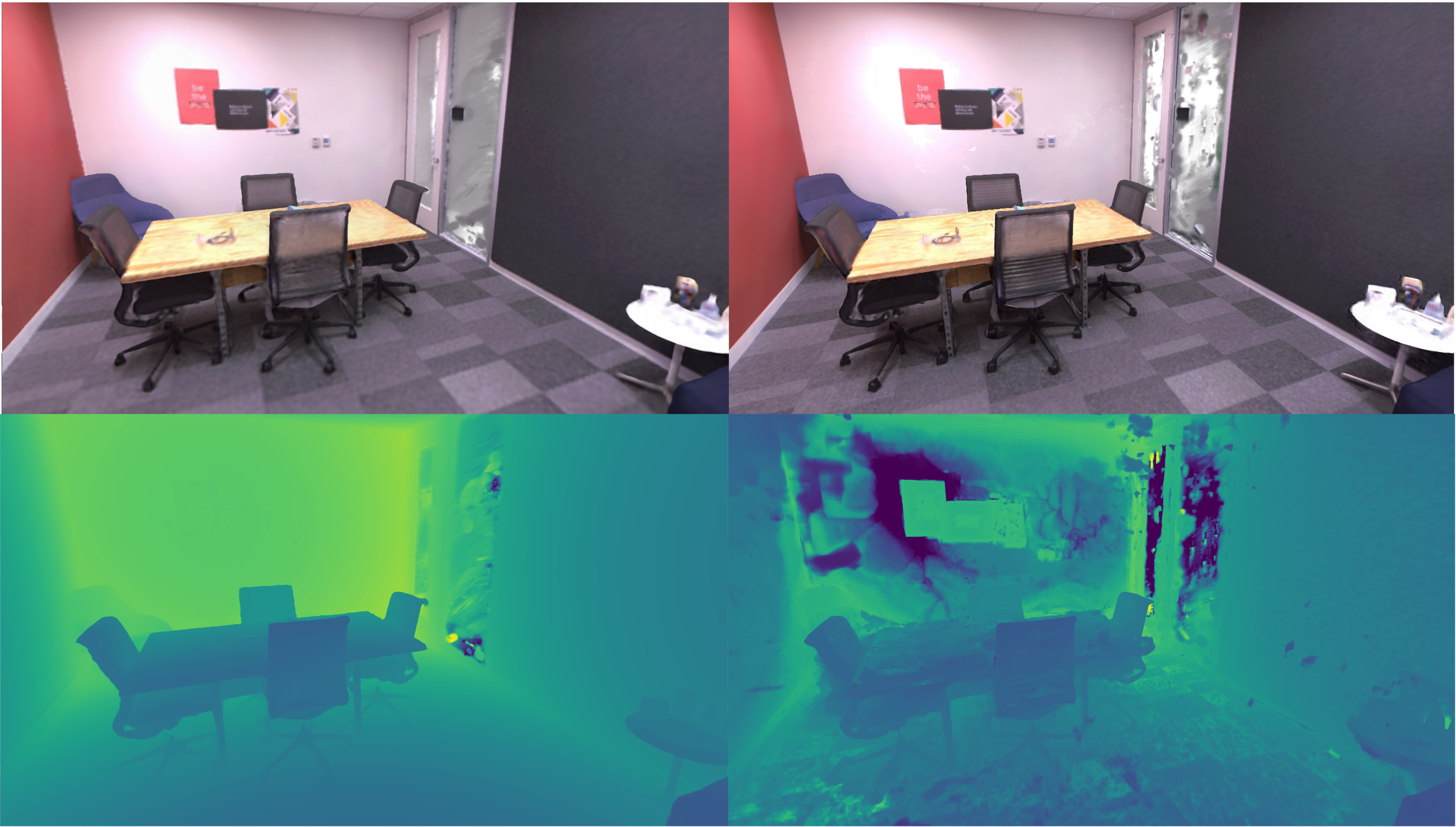}
   \caption{Color and depth rendering from two different radiance fields trained on office-2 in the Replica dataset: One trained with an RGB-D sensor(left column) and the other one trained with only an RGB sensor(right column), both trained on a large set of images viewing the scene from several different angles.}
   \label{fig:rgb_vs_rgbd}
\end{figure}

\begin{table}
  \centering
  \resizebox{0.7\hsize}{!}{
  \begin{tabular}{@{}lcc@{}}
    \toprule
     & RGB PSNR$\uparrow$ & Depth diff. (m/pixel)$\downarrow$\\
    \midrule
    RGB   & 30.612    & 0.6971 \\
    RGB-D & 28.570    & 0.0090 \\
    \bottomrule
  \end{tabular}}
  \caption{Comparison of radiance field color and geometry accuracy, including RGB and RGB-D sensor for training on Office-2 sequence of Replica dataset.}
  \label{tab:compareRGB_RGBD}
\end{table}

\section{Novel View Synthesis vs 3D Reconstruction}
\label{sec:nerfs_for_xr_and_robotics}
\begin{figure*}[h]
  \centering
   \includegraphics[width=\textwidth]{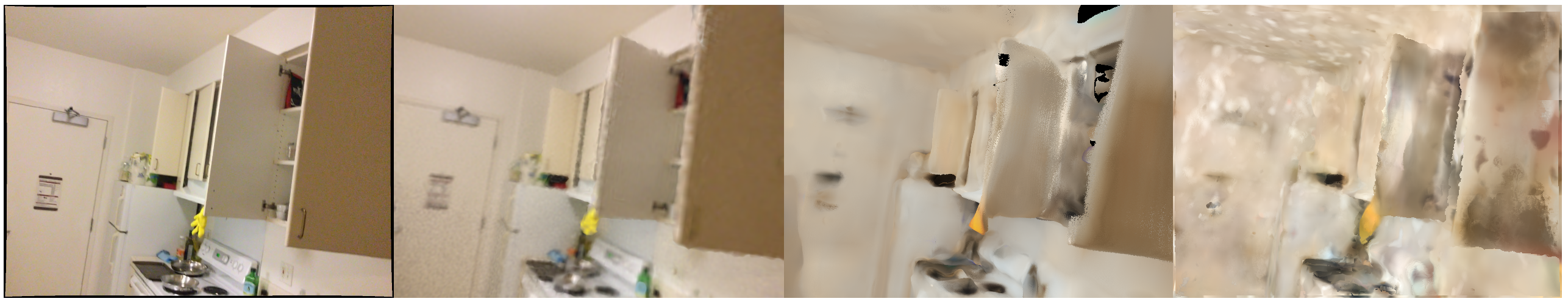}
   \caption{Rendered RGB images on the same camera position from trained radiance field under ScanNet scene0207. From left to right, the order is as follows: ground truth RGB image, from ours, from Vox-
   Fusion, and from NICE-SLAM.}
   \label{fig:scannet_compare_rgb}
\end{figure*}

We want to emphasize the difference between a model for novel view synthesis and a 3D reconstruction. A 3D reconstruction is a task that aims at recreating an accurate representation of the geometry of the target scene. View synthesis, on the other hand, is used to recreate an accurate representation of the appearance of the target scene. While these two statements sound similar, they are not the same. In their current form \cite{mildenhall2021nerf, barron2021mip, fridovich2022plenoxels} novel view synthesis models are trained based on RGB data from a limited number of views and will only produce reasonable novel appearance renderings within the distribution of camera poses that were used for training the model. In the case of an image rendered from a camera pose outside of the training pose distribution, the image will most likely not correspond to the expected result. This is because the underlying geometry from a series of images is ambiguous for areas of the images with no gradients, something that has been well-known in the photogrammetry community for decades. However, although the spatial geometry of such a region is unknown, the appearance of the same image region is known, at least locally, in the training pose distribution, allowing for the production of a model that can produce highly convincing images with incorrect underlying geometry. 

This is exactly the reason why radiance models can be so deceiving, but it also gives us a glimpse of our own faulty assumptions when determining scene geometry. If we are presented with an image from a novel view synthesis model that produces an accurate appearance but contains an incorrect underlying geometry, we will not perceive the incorrect model geometry but rather assume a geometry closer to the true scene geometry because that intuitively makes more sense. A good example of this can be seen in \cref{fig:rgb_vs_rgbd}.
Although the model trained with the RGB sensor looks good in the color rendering, we can see that the underlying geometry is very different from what we would expect compared to the radiance field trained with the RGB-D sensor.


The problem of ambiguous geometry can be mitigated either by training on a scene with color gradients covering significant parts of the scene, thereby reducing the total possible ambiguity, or by training the model with training images covering a 360-degree sphere around the model (inward-facing scene), effectively performing space carving \cite{kutulakos2000theory}. This will encourage the model to approach the true scene geometry. For many practical applications like XR and robotics, these mitigation strategies are often infeasible and/or impossible. Consider, for instance, an example of a room with monocolored walls (outward-facing scene). There will be little to no color gradient on the walls, and capturing all objects in the room from every angle places significant requirements on the data capture process.

However, if a depth sensor is adopted, this problem can be eradicated, as this allows for the estimation of a geometrically correct model with just a very limited coverage of the scene. A second, less ideal solution is to adopt an assumption on the geometry. For instance, assume areas with no color gradient are smooth, planar \cite{newcombe2011dtam}, or other more advanced solutions like inference from a pre-trained neural network \cite{eftekhar2021omnidata}. This would reduce the ambiguity of the geometry but also induce a bias that might either help or hinder the 3D reconstruction process depending on the assumption's correctness for any particular scene. 

The RGB-D sensor is becoming more ubiquitous, providing an unbiased representation of the geometry and allowing for easy integration. Additionally, the depth signal is often a lower frequency signal than the color signal, and therefore, it also helps increase the radius of convergence when using dense image-to-model alignment.  For these two reasons, we focus on the use of the RGB-D sensor in this work.


\section{Volumetric Rendering Basics}
All radiance field learning approaches share the volumetric rendering equations initially described in \cite{kajiya1984ray} and discretized in \cite{mildenhall2021nerf}. They explain how a pixel color is rendered based on a volume containing continuously valued implicit density and color functions. 
\\

\noindent \textbf{Color rendering.} Let $\hat{C}$ denote the rendered RGB color value of a single pixel from the radiance field model. $\hat{C}$ is obtained by accumulating $N$ rendered sample values computed by the density $\sigma_i$ and color values $c_i$ for sample points $\vec{p_i}$ along the ray $\vec{r} = \vec{o} + t_i \cdot \vec{d}$, $i \in [0,1,...,N]$. $\vec{o}$ is the camera center, $\vec{d}$ is the ray direction from the camera center to the pixel in the image plane and $t_i$ is the distance from the camera center. The distances between samples are denoted as $\delta_i$. The discrete rendering equations can then be expressed as follows:

\begin{equation}
    \label{eq:volume_rendering_color}
    \hat{C}(\vec{\sigma},\vec{c}) = \sum_{i=1}^N T_i (1-\exp{(-\sigma_i \delta_i)})c_i,
\end{equation}

where,
\begin{equation}
    \label{eq:transmittance}
    T_i = \exp{\left( -\sum_{j=1}^{i-1} \sigma_j \delta_j \right)}.
\end{equation}
The value $T_i$ represents the amount of light transmitted to sample $i$ along the ray $\vec{r}$. This is essentially the remaining light of a quota that is not reflected by all previous samples $j = [1,..,i-1]$. \\

\noindent \textbf{Depth rendering.} Let $\hat{D}$ denote the expected depth value of a single pixel rendered from the radiance field model. Depth $\hat{D}$ can be estimated analogously to color rendering by treating the sample distance $t_i$ from the camera as the sample color value $c_i$ in \cref{eq:volume_rendering_color}, resulting in the following:

\begin{equation}
    \label{eq:volume_rendering_depth}
    \hat{D}(\vec{\sigma}) = \sum_{i=1}^N T_i(\vec{\sigma}) (1-\exp{(-\sigma_i \delta_i)})t_i,
\end{equation}

with $T_i(\vec{\sigma})$ as defined in \cref{eq:transmittance}.




\section{Method}
\label{sec:method}
\subsection{Overview}
Our proposed algorithm is based on the voxel grid representation for radiance field optimization \cite{yu2021plenoctrees, fridovich2022plenoxels} and is divided into two separate parts: One offline mapping algorithm using RGB-D data with known poses to create a model of the scene, and one online tracking algorithm using the map in an image-to-model alignment scheme for pose optimization. All the partial derivatives of the loss function needed for both tasks are analytically calculated and implemented in CUDA for fast processing on GPUs.


\subsection{Mapping}
\begin{figure}[h]
  \centering
   \includegraphics[width=0.9\linewidth]{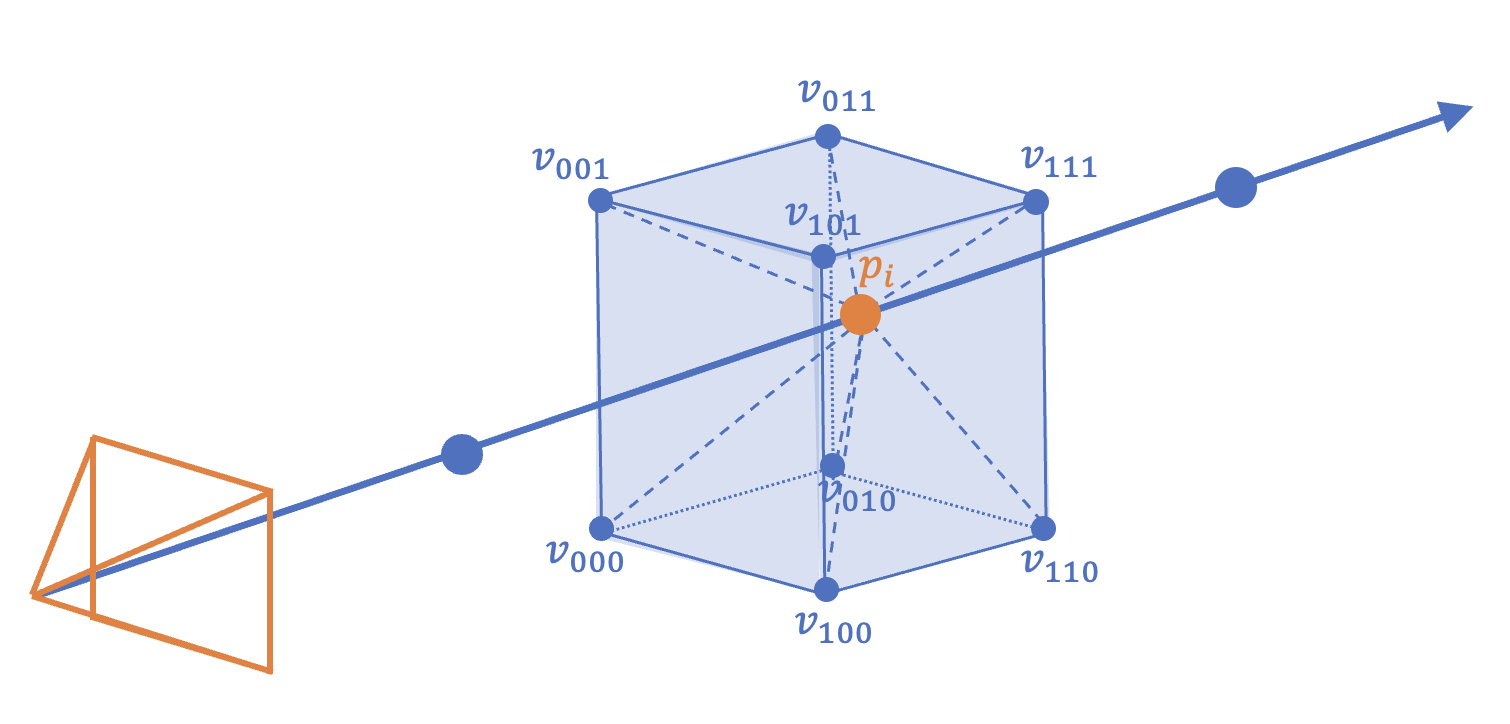}
   \caption{Illustration of trilinear interpolation on sample point $p_i$ lying on the sample ray in a voxel base radiance field.}
   \label{fig:tri-lerp}
\end{figure}

\noindent \textbf{Model representation.} Instead of using a neural network as a model representation as proposed in \cite{mildenhall2021nerf}, we utilize the Plenoxel representation \cite{fridovich2022plenoxels}, which represents the radiance field as a sparse voxel grid utilizing trilinear interpolation to produce a continuously valued implicit density and color functions, see \cref{fig:tri-lerp}. Each vertex in the voxel grid is represented by 28 scalar values: one density value and nine values per color channel. The multiple values per color channel make up the coefficients of spherical harmonics functions, allowing for modeling view dependencies caused by Lambertian surfaces and specular reflections.
The main advantage of using a voxel grid compared to a single neural network is the significant reduction in computational effort needed in both training and inference of the model. 
Practically the direct data access of the voxel grid representation reduces the training time by two orders of magnitude and allows for real-time image rendering \cite{fridovich2022plenoxels}. \\ 

\noindent \textbf{RGB-D mapping.} Mapping based on RGB-D data involves simultaneous optimization of the radiance field using the RGB-D sensor color values $C$ and depth value $D$ from the training images with known poses for supervision. The photometric color loss $\mathcal{L}_p$ and geometric depth loss $\mathcal{L}_g$ are defined as follows:

\begin{equation}
    \mathcal{L}_{p}(\hat{C}(\sigma, c)) = \frac{1}{M}\sum_{i=1}^M{||\hat{C}(\sigma_i,c_i) - C||^2},
    \label{eq:loss_color}
\end{equation}

\begin{equation}
    \mathcal{L}_{g}(\hat{D}(\sigma)) = \frac{1}{M}\sum_{i=1}^M{||\hat{D}(\sigma) - D||^2}.
    \label{eq:loss_depth}
\end{equation}

The mapping is performed by non-linear optimization based on the differentiation of \cref{eq:volume_rendering_color} and \cref{eq:volume_rendering_depth} with respect to all color $\vec{c}$ and density $\vec{\sigma}$ values. 

Partial derivatives of the photometric loss ${L}_{p}$, and geometric loss ${L}_{g}$ with respect to the variables $(\sigma_i, c_i)$ gives us:
\begin{equation}
    \frac{\partial \mathcal{L}_{p}(\hat{C}(\sigma, c))}{\partial (\sigma, c)} = \frac{\partial \mathcal{L}_p(\hat{C})}{\partial \hat{C}}\left({\frac{\partial \hat{C}(\sigma, c)}{\partial c_i} 
    + \frac{\partial \hat{C}(\sigma, c)}{\partial \sigma_i}}\right),
\end{equation}

\begin{equation}
    \frac{\partial \mathcal{L}_{g}(\hat{D}(\sigma))}{\partial (\sigma)} = \frac{\partial \mathcal{L}_p(\hat{D})}{\partial \hat{D}}{\frac{\partial \hat{D}(\sigma)}{\partial \sigma}}.
\end{equation}

The derivatives of the rendered color $\hat{C}$ are calculated in \cite{yu2021plenoctrees}, but for the sake of completeness, we include them here:

    
\begin{equation}
    \label{eq:derivative_color_wrt_color}
    \frac{\partial \hat{C}}{\partial c_i} (\sigma, c) = w_i(\sigma) = T_{i}(1 - \exp(-\sigma_i\delta_i)),
\end{equation}
\begin{equation}
    \label{eq:derivative_color_wrt_density}
\resizebox{.9\hsize}{!}{$
    \frac{\partial \hat{C}}{\partial \sigma_i} (\sigma, c) = \delta_{i} \left[ c_{i}T_{i+1}(\sigma) - \hat{C} + \sum_{j=0}^{i} c_jw_j(\sigma)\right]
    $}
\end{equation}
Based on the color derivative, we derive the depth derivative $\hat{D}$ in an analogous fashion. By interpreting the distance value $t_i$ as the color value $c_i$, we can re-write the equation \cref{eq:derivative_color_wrt_density} for the depth rendering equation: 
\begin{equation}
    \label{eq:derivative_depth_wrt_density}
    \frac{\partial \hat{D}}{\partial \sigma_i} (\sigma) = 
    \delta_{i} \left[ t_{i}T_{i+1}(\sigma) - \hat{D} + \sum_{j=0}^{i} t_jw_j(\sigma)\right]
\end{equation}

As the partial derivatives for equations are all analytically defined, it allows us to implement them directly in a custom CUDA kernel. Building on the code developed for Plenoxel \cite{fridovich2022plenoxels} to boost computational speed.

The final loss function we employ during mapping is represented as follows:
\begin{equation}
    \mathcal{L} = \mathcal{L}_p + \lambda_d\mathcal{L}_{g},
    \label{eq:tot_loss}
\end{equation}
where $\lambda_d$ is the scaling factor of the geometric loss.


\subsection{Tracking}

Up until this point, we have assumed known and fixed camera poses while treating the model parameters $(\vec{\sigma}, \vec{c})$ as variables. To estimate pose on a model build apriori, we use the same volumetric equations but consider the pose as the variable and the model parameters as static. Our objective is then to perform image-to-model alignment using the fixed volumetric color and density parameters from the radiance field, referencing the RGB-D input.
To achieve this, we need to determine the camera pose by finding the derivative of the color and depth rendering with respect to the ray $\vec{r} = \vec{o} + t_i \vec{d}$. 
We do this by leveraging the use of an alternative version of the chain rule:
 \begin{align}
    \begin{split}
        \frac{\partial \hat{C}}{\partial \vec{r}} &= \sum_{i=1}^{N} \frac{\partial \hat{C}}{\partial \vec{p}_i} \frac{\partial \vec{p}_i}{\partial \vec{r}} 
        \\&= \sum_{i=i}^{N} \left( \frac{\partial \hat{C}}{\partial \sigma_i} \frac{\partial \sigma_i}{\partial \vec{p}_i} + \frac{\partial \hat{C}}{\partial c_i} \frac{\partial c_i}{\partial \vec{p}_i}\right) \frac{\partial \vec{p}_i}{\partial \vec{r}}
    \label{eq:derivative_color_wrt_ray}
    \end{split}
    \\
    \begin{split}
        \frac{\partial \hat{D}}{\partial \Vec{r}} &= \sum_{i=i}^{N} \frac{\partial \hat{D}}{\partial \sigma_i} \frac{\partial \sigma_i}{\partial \Vec{p}_i} \frac{\partial \Vec{p}_i}{\partial \Vec{r}}.
        \label{eq:derivative_depth_wrt_ray}
    \end{split}
\end{align}
The partial derivatives $\frac{\partial \hat{C}}{\partial \sigma_i}$, $\frac{\partial \hat{C}}{\partial c_i}$, $\frac{\partial \hat{D}}{\partial \sigma_i}$ are already given in equations \ref{eq:derivative_color_wrt_color}, \ref{eq:derivative_color_wrt_density} and \ref{eq:derivative_depth_wrt_density} and are shared by all volumetric radiance field methods based on equations \ref{eq:volume_rendering_color} and \ref{eq:volume_rendering_depth}, while $\frac{\partial \sigma_i}{\partial \vec{p}_i}$, $\frac{\partial c_i}{\partial \vec{p}_i}$ and $\frac{\partial \sigma_i}{\partial \Vec{p}_i}$ are model specific. For our chosen representation based on the tri-linear interpolation functions, these partial derivatives produce quite messy functions, so we reserve the complete analytical equations for Appendix A. 

The derivative of a sample point $\vec{p_i}$ with respect to the ray $\vec{r}$ can further be broken down as:
\begin{table*}[ht]
  \centering
  \resizebox{\textwidth}{!}{%
  \begin{tabular}{@{}llcccccccccc@{}}
    \toprule
     Methods & Metric & Room-0 & Room-1 & Room-2 & Office-0 & Office-1 & Office-2 & Office-3 & Office-4 & Avg. & time(avg.)(ms)\\
    \midrule
    \multirow{3}{*}{Vox-Fusion~\cite{yang2022vox}} & \textbf{ATE}[m]$\downarrow$ 
    & 0.0042 & 0.0036 & 0.0090 & 0.6539 & 0.0029 & 0.0038 & 0.0042 & 0.0046 & 0.0046 & 
    \multirow{3}{*}{581.35} \\
  & \textbf{$\text{RPE}_t$}[m]$\downarrow$ 
  & 0.00391 & \textbf{0.0039} & 0.0088 & 0.0500 & 0.0032 & 0.0036 & 0.0039 & 0.0050 & 0.0046 \\
  & \textbf{$\text{RPE}_r$}[$^{\circ}$]$\downarrow$ 
  & 0.1076 & 0.1275 & 0.4034 & 2.6541 & 0.1501 & 0.1331 & 0.1316 & 0.1471 & 0.1715 \\
  \midrule
  \multirow{3}{*}{$\text{Ours}^1$} & \textbf{ATE}[m]$\downarrow$ 
  & \textbf{0.0017} & 0.0036 & \textbf{0.0020} & \textbf{0.0077} & \textbf{0.0020} & \textbf{0.0027} & \textbf{0.0022} & \textbf{0.0027} & \textbf{0.0031} 
  & \multirow{3}{*}{419.86} \\
  & \textbf{$\text{RPE}_t$}[m]$\downarrow$ 
  & \textbf{0.0008} & 0.0059 & \textbf{0.0013} & \textbf{0.0034} & \textbf{0.0012} & \textbf{0.0009} & \textbf{0.0008} & \textbf{0.0011} & \textbf{0.0019} \\
  & \textbf{$\text{RPE}_r$}[$^{\circ}$]$\downarrow$ 
  & \textbf{0.0191} & \textbf{0.0736} & \textbf{0.0483} & \textbf{0.1564} & \textbf{0.0440} & \textbf{0.0282} & \textbf{0.0199} & \textbf{0.0278} & \textbf{0.0522} \\
  \midrule
  \midrule
  \multirow{3}{*}{NICE-SLAM~\cite{zhu2022nice}} & \textbf{ATE}[m]$\downarrow$ 
  & 0.0119 & 0.0220 & 0.0334 & \textbf{0.0100} & 0.0042 & 0.0093 & 0.0862 & 0.0573 & 0.0293 & \multirow{3}{*}{149.57} \\
  & \textbf{$\text{RPE}_t$}[m]$\downarrow$ 
  & 0.0170 & 0.0229 & 0.0215 & 0.0143 & 0.0059 & 0.0131 & 0.0445 & 0.0305 & 0.0212 \\
  & \textbf{$\text{RPE}_r$}[$^{\circ}$]$\downarrow$ 
  & 0.3267 & 0.5103 & 1.4032 & 0.2700 & 0.2037 & 0.3247 & 1.0076 & 0.5058 & 0.5690 \\
  \midrule
  \multirow{3}{*}{$\text{Ours}^2$} & \textbf{ATE}[m]$\downarrow$ & 
  \textbf{0.0093} & \textbf{0.0061} & \textbf{0.0179} & 0.0117 & \textbf{0.0027} & \textbf{0.0056} & \textbf{0.0072} & \textbf{0.0061} & \textbf{0.0083} & \multirow{3}{*}{145.20}\\
  & \textbf{$\text{RPE}_t$}[m]$\downarrow$ 
  & \textbf{0.0065} & \textbf{0.0090} & \textbf{0.0108} & \textbf{0.0055} & \textbf{0.0025} & \textbf{0.0033} & \textbf{0.0046} & \textbf{0.0039} & \textbf{0.0058} \\
  & \textbf{$\text{RPE}_r$}[$^{\circ}$]$\downarrow$ 
  & \textbf{0.1494} & \textbf{0.1903} & \textbf{0.4833} & \textbf{0.2034} & \textbf{0.1196} & \textbf{0.1116} & \textbf{0.1201} & \textbf{0.1113} & \textbf{0.1861} \\
    \bottomrule
  \end{tabular}
  }
  \caption{Trajectory estimation results on the Replica dataset. The lowest errors are indicated in bold. Time here stands for tracking time per frame. We provide the results for two different configurations of our method, varying in number of samples per frame.}
  \label{tab:replica_ate}
\end{table*}

\begin{equation}
    \frac{\partial \Vec{p}_i}{\partial \Vec{r}} = \left[\frac{\partial \Vec{p}_i}{\partial \Vec{o}}, \frac{\partial \Vec{p}_i}{\partial \Vec{d}}\right] = \left[1, t_i\right].
\end{equation}

This allows for direct optimization of the position and orientation of the camera parameters.


\subsection{Optimization Details}
The optimization heuristics are kept as simple as possible, and mapping is done using a random set of sample rays from all available images, while tracking is done by randomly sampling rays from the current image. The optimizers used are RMSProp \cite{hinton2012neural} for the mapping, and Adam \cite{kingma2014adam} for the mapping. There is no importance sampling, keyframing, or similar optimization schemes.



\section{Experiments}
\label{sec:experiment}
\subsection{Experimental Setup}
We test both mapping and tracking algorithms on both synthetic and real indoor datasets of varying sizes and compare them to existing state-of-the-art algorithms performing similar tasks. \\

\noindent \textbf{Datasets.} 
Two different RGB-D datasets are used: 1) The Replica dataset \cite{straub2019replica} containing a total of 18 different synthetic indoor environments with highly accurate sensor data, 2) The ScanNet dataset \cite{dai2017scannet}, a real-world dataset also captured from indoor scenes, containing over 1000 unique sequences. As a real-world dataset, the latter inherently includes a significant amount of missing and imperfect depth measurements. 
We use a subset of eight sequences from the Replica dataset and five sequences from the ScanNet dataset. Both of these subsets have been the standard for comparison in previous works \cite{yang2022vox, zhu2022nice, sucar2021imap}. \\

\begin{table}[t]
  \centering
  \resizebox{0.9\hsize}{!}{
  \begin{tabular}{@{}lcc@{}}
    \toprule
     & RGB PSNR$\uparrow$ & Depth diff. (m/pixel)$\downarrow$\\
    \midrule
    Vox-Fusion~\cite{yang2022vox}   & 19.379 & 0.0705 \\
    NICE-SLAM~\cite{zhu2022nice} & 18.455 & 0.0514 \\
    Ours & 24.411 & 0.0469 \\
    \bottomrule
  \end{tabular}}
  \caption{comparison of map geometry accuracy on ScanNet. The values are average through 5 sequences reported in \cref{tab:traj_result_scannet}.}
  \label{tab:compareMapping}
\end{table}
\begin{table*}[ht]
  \caption{Trajectory estimation results on the ScanNet dataset. The lowest errors are indicated in bold. Time here stands for tracking time per frame. Both NICE-SLAM and ORB-SLAM2 failed to localize camera pose in the middle of sequence 0181. The average errors of these methods were computed excluding frame 0181 and are indicated with (*) marks.}
  \centering
  \resizebox{0.8\textwidth}{!}{%
  \begin{tabular}{@{}llccccccc@{}}
  \toprule
  Methods & Metric & 0000 & 0106 & 0169 & 0181 & 0207 & Avg. & time(avg.)(ms)\\
  \midrule 
  \multirow{3}{*}{Vox-Fusion~\cite{yang2022vox}}
  & \textbf{ATE}[m]$\downarrow$ 
  & 0.0274 & 0.2424 & 0.0315 & 0.0924 & 0.0323 & 0.0852 & \multirow{3}{*}{1021.19} \\
  & \textbf{$\text{RPE}_t$}[m]$\downarrow$ 
  & 0.0222 & 0.0763 & 0.0284 & 0.0443 & 0.0303 & 0.0403 \\
  & \textbf{$\text{RPE}_r$}[$^{\circ}$]$\downarrow$ 
  & 0.6157 & 2.7116 & 0.8805 & 1.8274 & 1.0733 & 1.4217 
\\ 
  \midrule
  \multirow{3}{*}{NICE-SLAM~\cite{zhu2022nice}}
  & \textbf{ATE}[m]$\downarrow$ 
  & 0.0405 & 0.1188 & 0.1952 & - & 0.0491 & 0.1009* &\multirow{3}{*}{360.70}  \\
  & \textbf{$\text{RPE}_t$}[m]$\downarrow$ 
  & 0.0360 & 0.0666 & 0.0643 & - & 0.0457 & 0.0531* \\
  & \textbf{$\text{RPE}_r$}[$^{\circ}$]$\downarrow$ 
  & 1.0862 & 2.3725 & 1.4703  & -  & 1.4871  & 1.6040* \\
  \midrule
  \multirow{3}{*}{Ours} & \textbf{ATE}[m]$\downarrow$ 
  & \textbf{0.0246} & \textbf{0.0387} & \textbf{0.0165} & \textbf{0.0373} & \textbf{0.0259} & \textbf{0.0286}  &\multirow{3}{*}{350.84}\\
  & \textbf{$\text{RPE}_t$}[m]$\downarrow$ 
  & \textbf{0.0177} & \textbf{0.0209} & \textbf{0.0160} & \textbf{0.0264} & \textbf{0.0230} & \textbf{0.0208} \\
  & \textbf{$\text{RPE}_r$}[$^{\circ}$]$\downarrow$ 
  & \textbf{0.5030} & \textbf{0.6216} & \textbf{0.4484} & \textbf{0.7872} & \textbf{0.7231} & \textbf{0.6166}\\
  \midrule
  \multirow{3}{*}{ORB-SLAM2~\cite{mur2017orb}}
  & \textbf{ATE}[m]$\downarrow$ 
  & 0.0779 & 0.0838 & 0.1038 & - & 0.0898 & 0.0888* &\multirow{3}{*}{36.30}  \\
  & \textbf{$\text{RPE}_t$}[m]$\downarrow$ 
  & 0.0933 & 0.0516 & 0.0625 & - & 0.0848 & 0.0730* \\
  & \textbf{$\text{RPE}_r$}[$^{\circ}$]$\downarrow$ 
  & 2.8584 & 2.9895 & 2.3596  & -  & 3.5897  & 2.9493* \\
  \bottomrule
  \end{tabular}}%
  \label{tab:traj_result_scannet}
\end{table*}

\noindent \textbf{Comparison algorithms.} To the best of our knowledge, no other algorithms are currently performing asynchronous mapping and tracking in a radiance field. Therefore, to draw comparisons with existing approaches, we select NICE-SLAM \cite{zhu2022nice}, and Vox-Fusion \cite{yang2022vox} as competing methods due to their status as state-of-the-art algorithms for radiance field-based simultaneous localization and mapping. To make the comparison fair, we modify these algorithms first to optimize the map using ground truth poses with a subset of the sequence images and then perform tracking based on that model. \\

\noindent \textbf{Metric.} For the offline mapping task, we report on two primary metrics to illustrate the geometrical accuracy of the constructed map: the average L1 depth loss in metric units and PSNR score of the RGB values of the model. These metrics are calculated based on randomly sampled pixels from randomly sampled images from the sequence. 

To measure tracking accuracy, we use three key metrics stemming from the root mean square error (RMSE): Absolute Trajectory Error (ATE)\cite{sturm2012benchmark}, and the Relative Pose Error for both translation ($\text{RPE}_t$) and rotation ($\text{RPE}_r$). 
ATE reflects the global accuracy of the trajectory as it is sensitive to drifting over time. We put a greater emphasis on the $\text{RPE}_t$ and $\text{RPE}_r$ due to local accuracy being a more relevant metric when tracking based on a map known apriori.
We measure RPE in 1-meter intervals. These tracking metrics were implemented using the "evo" Python package for odometry and SLAM evaluation \cite{Grupp2017evo}.\\

\noindent \textbf{Implementation details and parameter selection.} All experiments were conducted on a machine equipped with an Intel Core i9-11900KF CPU and an NVIDIA RTX 3090 graphics card, with custom CUDA code for efficient runtimes. 
For all experiments, our method used a dynamic voxel grid with a peak voxel resolution of $512^3$, irrespective of the size of the scene.
While all images are used for tracking, only every 10th image is used for mapping. Pixels with invalid depth measurements are disregarded in both mapping and tracking.

For mapping, we allot our algorithm roughly 6 seconds per frame for mapping, while the competing methods are given up to 18 seconds per frame. The parameters of the competing methods are adjusted to take full advantage of this extra time.

For tracking, the number of sample rays and the number of iterations varied between each method and each dataset. We adhere to the default configurations for these parameters provided by NICE-SLAM and Vox-Fusion for each dataset, while we adjusted the parameters for our model to match the per-frame tracking time of the two comparison models.


\subsection{Mapping Results}

\noindent \textbf{RGB vs RGB-D mapping.} To first demonstrate the point highlighted in \cref{sec:nerfs_for_xr_and_robotics}, we test the difference in mapping performance between the original Plenoxel and our RGB-D modified mapping algorithm on the Office-1 scene of the Replica dataset. The results are shown in \cref{fig:rgb_vs_rgbd} and \cref{tab:compareRGB_RGBD}.
Despite the PSNR for RGB rendering actually being higher for the RGB-only reconstruction, the average L1 depth error per pixel is 70cm, compared to the RGB-D L1 depth error of just 9mm. Confirming that although RGB alone might look convincing for use in novel view synthesis, it is insufficient for learning the underlying geometrical information, especially in areas with weaker gradients.\\

\noindent \textbf{Baseline comparison.} The results of computing the average PSNR and L1 difference for five sequences of ScanNet are presented in \cref{tab:compareMapping}. Even with significantly less training time than the competing methods, our model achieves a significantly better RGB PSNR and produces the best 3D reconstruction by exhibiting the smallest L1 depth loss, keeping within a range of 5cm. 
The qualitative difference in RGB estimation can be observed in \cref{fig:scannet_compare_rgb}, where our method shows a visually more correct image compared to the over-smoothed results from baselines. This builds a strong foundation for achieving better tracking accuracy.



\subsection{Tracking Results}
\noindent \textbf{Replica dataset.} For the replica dataset, we report two different sampling configurations of our system. One that roughly matches the processing time of Vox-Fusion, and one that matches NICE-SLAM. \cref{tab:replica_ate} shows the estimated pose tracking accuracy for the entire sequence on the Replica dataset. NICE-SLAM estimates the pose with efficient use of samples, with relatively few rays. This leads to a much faster processing speed than Vox-Fusion but still with a comparative reduction in overall accuracy. When we matched our method to NICE-SLAM's tracking speed, our method outperformed NICE-SLAM across almost all metrics and sequences. 
Notably, NICE-SLAM showed instability in the rotational aspect of relative pose accuracy, as it relied on neural network processing with a relatively large voxel size, leading to challenges in precisely estimating subtle rotational differences. 
NICE-SLAM marginally outperformed our method in the Office-0 sequence in terms of ATE but is beaten on the relative pose metrics. 
Vox-Fusion showed remarkable tracking performance, taking advantage of its considerably longer tracking time. It trailed only slightly behind our proposed method in most of the sequences. 
However, like NICE-SLAM, it also tended to have lower accuracy for the orientation of the relative pose.
For an analysis of our model's speed vs tracking accuracy trade-off, please see Appendix B.\\

\noindent \textbf{ScanNet dataset.} \cref{tab:traj_result_scannet} displays the comparison of tracking performance on the five ScanNet sequences. 
Our method excels in all metrics across all sequences when compared to the comparison algorithms while demanding significantly less computation time. Our method even outperforms Vox-Fusion, using only a third of the computation time. Since our method relies on directly optimizing small voxels, we initially presumed the significant amount of invalid depth measurements in ScanNet data might lead to holes in the final model, but our method successfully filled these gaps and created a seamless, hole-free model.

To ground our work with pre-NeRF research, we additionally provide comparisons with a more traditional RGB-D SLAM method that does not use a radiance field model, ORB-SLAM2\cite{mur2017orb}, which has long been the standard baseline in SLAM systems. 
We do not separate the mapping and tracking tasks for ORB-SLAM2 as we have done for the other methods because we observed worse performance when doing this, but we have included it as a point of reference.
Although a feature-based system like ORB-SLAM2 produces good results in significantly less time than any of the other algorithms, it cannot beat the accuracy of any of the dense methods. This primarily comes down to three factors: 1) The offline mapping gives a major advantage to the dense systems, 2) the dense systems use all image information, and 3) the ScanNet sequences have a significant amount of repetitive patterns on carpets and walls that make the data correspondence task difficult for ORB-SLAM2.



\section{Conclusion}



We have presented the analytical augmentation of the Plenoxel algorithm to allow for RGB-D mapping and tracking 
based on the radiance field equations. We have also argued for and shown both qualitatively and quantitatively the need for RGB-D sensors to accurately and reliably reconstruct outward-facing scenes when modeling using methods from the current paradigm of learning-based radiance field algorithms. Our method achieves superior results with much less time on both the mapping and tracking tasks compared to state-of-the-art radiance field-based SLAM methods when modified to perform offline mapping based on ground truth pose data.

\section*{Acknowledgement}

This work is financially supported by the Korea Agency for Infrastructure Technology Advancement (KAIA) grant funded by the Ministry of Land, Infrastructure and Transport (Grant RS-2021-KA160637) and the Norwegian Research Council in the project Autonomous Robots for Ocean Sustainability (AROS), project number 304667.


{\small
\bibliographystyle{ieee_fullname}
\bibliography{egbib}
}

\newpage
\newpage
\newpage
\section*{APPENDIX}
\begin{figure}[h]
  \centering
   \includegraphics[width=\linewidth]{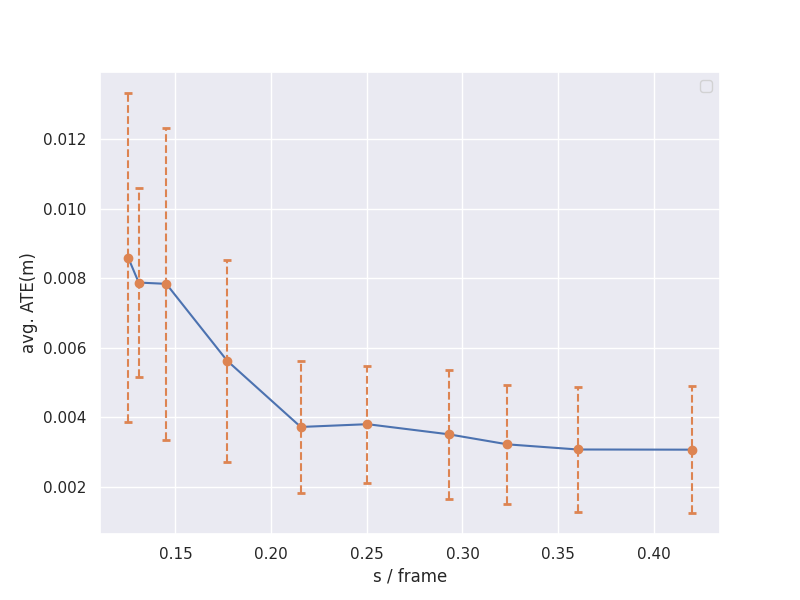}
   \caption{Average ATE error on Replica subsets regarding to the tracking speed. Standard deviation of error along 5 sequences are displayed in dotted line.}
   \label{fig:speed_analysis}
\end{figure}
\subsection*{Appendix A}
\label{App:A}
We explain the derivative of a tri-linear interpolated grid function with respect to a sample location as is present in equations 
8 and 9 from the paper
Let $p_i = (x_i, y_i, z_i)$ be the sample location and let the function $f(p)$ represent the tri-linearly interpolated grid function (Either $\vec{c}$ or $\vec{\sigma}$ in our case) where $[v_{000},...,v_{111}]$ are the eight closest vertices of $p_i$. Further let $(x_0, y_0, z_0)$ represent the lattice points below, and  $(x_1, y_1, z_1)$ represent the lattice points above the location  $(x_i, y_i, z_i)$. The trilinear interpolation can then be described by the equation:

\begin{align}
    \begin{split}
        &f(p_i) = f(x, y, z) = v_i \\ 
        &\approx a_0 + a_1 x_i + a_2 y_i + a_3 z_i + a_4 x y_i + a_5 x_i z_i + a_6 y_i z_i + a_7 x_i y_i z_i\\
        &\text{where} \\
        &\begin{bmatrix}
        1 & x_0 & y_0 & z_0 & x_0 y_0 & x_0 z_0 & y_0 z_0 & x_0 y_0 z_0 \\
        1 & x_1 & y_0 & z_0 & x_1 y_0 & x_1 z_0 & y_0 z_0 & x_1 y_0 z_0 \\
        1 & x_0 & y_1 & z_0 & x_0 y_1 & x_0 z_0 & y_1 z_0 & x_0 y_1 z_0 \\
        1 & x_1 & y_1 & z_0 & x_1 y_1 & x_1 z_0 & y_1 z_0 & x_1 y_1 z_0 \\
        1 & x_0 & y_0 & z_1 & x_0 y_0 & x_0 z_1 & y_0 z_1 & x_0 y_0 z_1 \\
        1 & x_1 & y_0 & z_1 & x_1 y_0 & x_1 z_1 & y_1 z_1 & x_1 y_0 z_1 \\
        1 & x_0 & y_1 & z_1 & x_0 y_1 & x_0 z_1 & y_1 z_1 & x_0 y_1 z_1 \\
        1 & x_1 & y_1 & z_1 & x_1 y_1 & x_1 z_1 & y_1 z_1 & x_1 y_1 z_1 
        \end{bmatrix}
        \begin{bmatrix}
            a_0\\ a_1 \\ a_2 \\ a_3 \\ a_4 \\ a_5 \\ a_6 \\ a_7
        \end{bmatrix} 
        =
        \begin{bmatrix}
            v_{000} \\ v_{001} \\ v_{010} \\ v_{011} \\ v_{100} \\ v_{101} \\ v_{110} \\ v_{111}
        \end{bmatrix}
    \end{split}
\end{align}
As all voxels are locally independent we can treat the lower lattice points $(x_0, y_0, z_0)$ as $(0,0,0)$ greatly simplifying the equations.

Then if the partial derivatives of these equations are computed with respect to $p_i = (x_i, y_i, z_i)$ we get:

\begin{align}
    \begin{split}
        \frac{\partial v_i}{\partial x_i} = a_1 + a_4 y_i + a_5 z_i + a_7 y_i z_i\\
        \frac{\partial v_i}{\partial y_i} = a_2 + a_4 x_i + a_6 z_i + a_7 x_i z_i\\
        \frac{\partial v_i}{\partial z_i} = a_3 + a_5 x_i + a_6 y_i + a_7 x_i y_i\\
    \end{split}
\end{align}
\\
\\

\subsection*{Appendix B}
\label{App:B}

\cref{fig:speed_analysis} displays the speed-accuracy trade-off curves obtained by testing different settings across the eight sequences from the Replica dataset. 
Comparing the result of NICE-SLAM in table \ref{tab:replica_ate}, it indicates that even if we reduce the allotted tracking time of our method to just 0.075s per frame, our method still outperforms NICE-SLAMs results attained using double the computation time.

\end{document}